\documentclass{article}

\usepackage{microtype}
\usepackage{graphicx}
\usepackage{amsmath}

\usepackage{subfigure}
\usepackage{comment}
\usepackage{booktabs} 
\usepackage{amssymb}
\usepackage{multirow}
\usepackage[multiple]{footmisc}
\usepackage[outline]{contour}
\usepackage{enumitem}
\usepackage{hyperref}
\usepackage{tablefootnote}

\usepackage[accepted]{icml2021}

\icmltitlerunning{Mitigating Sampling Bias and Improving Robustness in Active Learning}

\begin{document}

\twocolumn[
\icmltitle{Mitigating Sampling Bias and Improving Robustness in Active Learning}



\icmlsetsymbol{equal}{*}

\begin{icmlauthorlist}
\icmlauthor{Ranganath Krishnan}{org1}
\icmlauthor{Alok Sinha}{org3}
\icmlauthor{Nilesh Ahuja}{org2}
\icmlauthor{Mahesh Subedar}{org1}
\icmlauthor{Omesh Tickoo}{org1}
\icmlauthor{Ravi Iyer}{org1}
\end{icmlauthorlist}

\icmlaffiliation{org1}{Intel Labs, Oregon (USA)}
\icmlaffiliation{org2}{Intel Labs, California (USA)}
\icmlaffiliation{org3}{Intel Corporation, Bangalore (India)}

\icmlcorrespondingauthor{Ranganath Krishnan}{ranganath.krishnan@intel.com}

\icmlkeywords{Machine Learning, Active Learning, Contrastive Learning, Bias in machine learning, ICML}

\vskip 0.3in
]



\printAffiliationsAndNotice{}  

\begin{abstract}
This paper presents simple and efficient methods to mitigate sampling bias in active learning while achieving state-of-the-art accuracy and model robustness. 
We introduce supervised contrastive active learning by leveraging the contrastive loss for active learning under a supervised setting. 
We propose an unbiased query strategy that selects informative data samples of diverse feature representations with our methods: supervised contrastive active learning (SCAL) and deep feature modeling (DFM). We empirically demonstrate our proposed methods reduce sampling bias, achieve state-of-the-art accuracy and model calibration in an active learning setup with the query computation 26x faster than Bayesian active learning by disagreement and 11x faster than CoreSet. The proposed SCAL method outperforms by a big margin in robustness to dataset shift and out-of-distribution.
\end{abstract}
\section{Introduction}
\label{sec:introduction}

Deep learning relies on a large amount of labeled data for training the models. Data annotation is often very expensive and time-consuming.  It is challenging to obtain labels for large-scale datasets in complex tasks such as medical diagnostics~\cite{irvin2019chexpert} that require specific expertise for data labeling, and semantic segmentation~\cite{cordts2016cityscapes} that requires pixel-wise labeling. Active learning~\cite{settles2009active,lewis1994sequential} is a promising solution that allows the model to choose the most informative data samples from which it can learn while requesting a human annotator to label the carefully selected data based on some query strategy. While active learning enables cost-efficient labeling, it introduces sampling bias~\cite{dasgupta2008hierarchical, dasgupta2011two, farquhar2021statistical}. Bias in sample selection has been studied decades ago in the field of statistics~\cite{heckman1979sample, heckman1990varieties}. Sampling bias in deep neural network training can cause undesired behavior in the models when deployed in real-world~\cite{buolamwini2018gender,bhatt2020uncertainty} with respect to fairness, robustness and trustworthiness.

Deep active learning has been widely studied~\cite{gal2017deep,shen2017deep,sener2018active,beluch2018power,ducoffe2018adversarial,yoo2019learning,kirsch2019batchbald}, but previous works do not address the sampling bias, which is also observed by~\citet{farquhar2021statistical}. As samples are queried iteratively based on heuristics in active learning, the acquired samples may not be a proper representative of the underlying data distribution.  We address the sampling bias in active learning by proposing an unbiased query strategy that leverages the feature representations from the neural network. The recent advancements in contrastive learning~\cite{chen2020simple,chen2020big,he2020momentum,chen2020improved} have resulted in state-of-the-art performance in unsupervised representation learning. We extend the contrastive loss to active learning in a supervised setting to leverage the powerful feature representations and propose scoring functions for the query strategy. We propose query strategy that is devised to select data samples with diverse features while maintaining an equal representation of samples from each class in order to mitigate the sampling bias. 

The previous works in deep active learning have mainly focused on improving the model accuracy as samples are acquired. Once the deep neural networks are trained, they face challenges with dataset shift and out-of-distribution data~\cite{ovadia2019can,krishnan2020improving}. It is important for the models to be well-calibrated and robust under such distributional shift in real-world settings in order to be deployed in safety-critical applications. We study the robustness of the models trained from different query methods in active learning. Our main contributions include: 


\begin{itemize}[topsep=-0.5pt]
\itemsep-0.1em 
\item We propose simple and efficient methods to mitigate the sampling bias in active learning by leveraging the learnt feature representations.
\item We introduce supervised contrastive active learning (SCAL) by leveraging the contrastive loss and the query strategy relying on the cosine similarity score in feature embedding space
\item We evaluate the model calibration and robustness in active learning setup, demonstrating our method yield well-calibrated models that outperform in robustness to dataset shift and out-of-distribution.
\end{itemize}
We compare our methods with high-performing active learning methods including CoreSet~\cite{sener2018active}, Learning Loss~\cite{yoo2019learning} and Bayesian active learning by disagreement (BALD)~\cite{gal2017deep}. SCAL method is 26x faster than BALD and 11x faster than CoreSet in query computation, while retaining expected calibration error (ECE) within 3\% higher than BALD. SCAL also outperforms all the methods by a bigger margin (AUROC more than 10\% and ECE lesser than 12\%) under distributional shift.

\section{Background and Related work}
\label{sec:background}

\subsection{Active Learning}
\label{subsec:active_learning}
Active learning aims to learn from a small set of informative data samples, which are acquired from the huge unlabeled dataset, thus minimizing the data annotation cost.
The acquired data samples are labeled by an oracle (e.g. human annotator), which is used for training the model. In this framework, the models are allowed to select the data from which they can learn based on a query strategy. 
The query strategy evaluates the informativeness of data samples; some of the commonly used strategies include the  uncertainty-based approach~\cite{lewis1994sequential, tong2001support, beluch2018power,gal2017deep, yoo2019learning}, diversity-based approach~\cite{brinker2003incorporating,sener2018active,shui2020deep} and query-by-committee approach~\cite{seung1992query,burbidge2007active}. For example, the data samples with higher uncertainty estimates are considered to be most useful in uncertainty-based query strategy. We refer to~\cite{settles2009active} for an overview of earlier works in active learning. Active learning is used in real-world applications including medical imaging~\cite{hoi2006batch,yang2017suggestive} and robotics~\cite{chao2010transparent}. 
\subsection{Bias in machine learning}
\label{subsec:bias_ml}
Bias can get introduced at different stages of machine learning (ML)~\cite{torralba2011unbiased} from dataset creation to model training. Dataset bias in ML can result in unintended consequences including unfair practices in the downstream tasks. 
Dataset capture and labeling can introduce bias~\cite{tommasi2017deeper} resulting in poor categorization and generalization of the classes.
Sampling bias occurs when the observations in the available data are not representative of the true data distribution.
Representation bias \cite{suresh2019framework} arises due to the difference in the relative distributions of the classes in the training data as compared to the population being represented.
Models trained in the presence of bias could be unfair and exhibit unwanted bias towards one or more target classes~\cite{mehrabi2019survey,buolamwini2018gender}, for example on the under-represented group. 

\subsection{Contrastive learning}
\label{subsec:contrastive_learning}
Contrastive learning~\cite{chen2020simple,chen2020big,he2020momentum,chen2020improved} requires positive and negative samples in a mini-batch to bring the positive samples closer in the feature space while pushing the negative samples further apart. In the self-supervised setting, positive samples are selected by applying data augmentation~\cite{chen2020simple}, and the rest of the samples are assumed to be negative examples. A contrastive loss using cosine similarity is defined on the normalized feature vectors for one positive and remaining negative samples.~\citeauthor{khosla2020supervised} propose a generalization to contrastive loss by extending to a fully-supervised learning setting by leveraging the label information.

To the best of our knowledge, the work presented in this paper is the first to apply contrastive learning method in active learning setup and also the first work to evaluate the robustness to distributional shift of models resulting from different query strategies in active learning setting. 

\citet{farquhar2021statistical} correct the bias in active learning by introducing the risk estimators, which applies corrective weighting to the acquired samples.
~\citet{sener2018active} proposed core-set approach to chose diverse samples with the k-center-greedy method, but the search procedure is very expensive resulting in high query time at every active learning iteration as noted in~\cite{shui2020deep}. We address the sampling bias in active learning by proposing scoring functions that are simple, computationally less expensive and yet efficient to select balanced and diverse samples. Our unbiased query strategy is based on the per-cluster feature similarity score in the feature embedding space. 

\section{Methods}
\label{sec:methods}
In this section, we present our proposed methods namely: supervised contrastive active learning (SCAL) and deep feature modeling (DFM) to mitigate sampling bias in active learning and select samples with diverse feature representations. We begin with problem formulation as we will follow the same notations in the rest of the paper.

\textbf{Problem formulation}: Consider a multi-class classification problem with a limited data labeling budget for the deep neural network model with weights $\mathrm{w}$, $\mathrm{y}=f_{\mathrm{\mathrm{w}}}\left(\mathrm{x}\right)$. Let $\mathcal{D}_{\mathrm{U}}=\left\{\mathrm{x_{u}}\right\}_{u=1}^{N}$ represent large pool of unlabeled data. Initially a small set of M samples are randomly sampled from $\mathcal{D}_{\mathrm{U}}$ and annotated by an oracle (human annotator) to obtain initial labeled set $\mathcal{D}_\mathrm{L}^{1}=\left\{\left({x_\ell}, {y_\ell}\right)\right\}_{\ell=1}^{M}$, where 
${y_{\ell}} \in \small{\{c_1,c_2, \cdots, c_k}\}$ is the ground-truth class label in K-class classification problem. 
The labeled M samples are removed from the unlabeled pool and model is trained with $\mathcal{D}_\mathrm{L}^{1}$. The next batch of $\mathrm{M}$ samples for the next training iteration of the active-learning cycle are chosen from the remaining unlabeled data $\mathcal{D}_\mathrm{U}^1 = \mathcal{D}_\mathrm{U} \setminus \mathcal{D}_\mathrm{L}^{1}$. These are chosen based on an acquisition function $\mathcal{Q}(\mathcal{D}, f_{\mathrm{w}}, \mathcal{S})$ that evaluates the samples from dataset $\mathcal{D}$ on the network $f_{\mathrm{w}}$ and determines the indices $I_\mathrm{M}$ of the M samples that yield the highest scores on a scoring function $\mathcal{S}$. $\mathcal{S}$ depends on the particular method used and typically indicates the uncertainty in the prediction of a sample from $f_{\mathrm{w}}$. The query function $\mathcal{Q}$ returns, therefore, the most informative samples for the next active learning cycle. These samples are annotated by an oracle and added to existing labeled set $\mathcal{D_{\mathrm{L}}}^{1}$ to create $\mathcal{D_{\mathrm{L}}}^{2}=\mathcal{D_{\mathrm{L}}}^{1} \cup \left\{\left({x_\ell}, {y_\ell}\right)\right\}_{\ell \in I_\mathrm{M}}$, and simultaneously are removed from the unlabeled set to obtain $\mathcal{D}_\mathrm{U}^2 = \mathcal{D}_\mathrm{U} \setminus \mathcal{D_{\mathrm{L}}}^{2}$. Thus, a sequence of [$\mathcal{D_{\mathrm{L}}}^{1}, \mathcal{D_{\mathrm{L}}}^{2}, \mathcal{D_{\mathrm{L}}}^{3}, ....$] labeled sets of size [M, 2M, 3M, ....] samples and corresponding trained models [$f_{\mathrm{\mathrm{w}}}^{1}, f_{\mathrm{\mathrm{w}}}^{2}, f_{\mathrm{\mathrm{w}}}^{3}, .... $] are obtained from every iteration of active learning cycle. This cycle is repeated until a desired model performance is achieved or until the query budget is reached.


\subsection{Supervised Contrastive Active Learning (SCAL)}
\label{subsec:scal}
We extend the contrastive loss proposed in~\cite{khosla2020supervised} and follow the training methodology to active learning in a supervised setting. At every active learning iteration, the model is trained with newly acquired labeled data using the loss function given by Equation~\eqref{eqn:supconloss}.
\begin{equation}
    \label{eqn:supconloss}
    \mathcal{L}_{con}=\sum_{i \in I} \frac{-1}{|P(i)|} \sum_{p \in P(i)} \log \frac{\exp \left(\boldsymbol{z}_{i} \cdot \boldsymbol{z}_{p} / T\right)}{\sum_{a \in A(i)} \exp \left(\boldsymbol{z}_{i} \cdot \boldsymbol{z}_{a} / T\right)}
\end{equation}
$\boldsymbol{z}$ is the feature from the neural network, T is scalar temperature parameter, $i \in I$ is the index of sample in augmented batch, $P\mathrm{(i)}$ is the set of all positives corresponding to the index $i$ and $|P(i)|$ is its cardinality and $A(i) \in I \setminus \{i\}$. We take advantage of the capability of contrastive loss in learning better feature representations as it pulls together clusters of samples from the same class and pushes apart clusters of samples from different classes in the feature space. Based on the predicted class, we leverage the cosine similarity scores between the sample and corresponding training features of the samples belonging to the same class to choose unbiased informative samples from the unlabeled pool. From each cluster, we choose to select the samples with the least similarity score in feature space with respect to the training samples belonging to the same class as the prediction. Similar properties of contrastive loss have been leveraged in~\cite{tack2020csi} for novelty detection in an unsupervised setting but without the importance of per-class or per-cluster feature representation.

We define a cluster-based feature similarity score that gives importance to finding the useful samples within the same cluster to acquire an unbiased and balanced labeled set, while still picking samples with diverse features. The acquisition function $\mathcal{Q}$ is intended to obtain $\frac{M}{K}$samples from each cluster in feature embedding space with the least similarity score as given by Equation~\eqref{eqn:sim_score}. We refer to this method as supervised contrastive active learning (SCAL). 
\begin{equation}
\label{eqn:sim_score}
   \mathcal{S} \triangleq {score}(\mathrm{x}) := \max _{\ell} \frac{z\left({x}_{\ell} |  {y}_{\ell}=c_{k}\right)}{ \|z\left({x}_{\ell} |  {y}_{\ell}=c_{k}\right)\|} \cdot z(\mathrm{x} | \mathrm{y}=c_{k})
\end{equation}

We further use the same query strategy on a model trained with cross-entropy loss and we refer to the method as \textbf{FeatureSim} (Feature Similarity) in the results section~\ref{sec:results}.


\subsection{Deep Feature Modeling (DFM)}
\label{subsec:dfm}
\citet{ahuja2019probabilistic} proposed an approach for detecting out-of-distribution samples which involves learning class-conditional probability distributions to the deep-features, $z$, of a DNN. At test time, the log-likelihood score of a test feature with respect to the learnt distributions is used to discriminate in-distribution samples (high likelihood) from OOD samples (low likelihood). Prior to learning distribution, the dimension of the feature space is reduced by applying a set of class-conditional PCA (principal component analysis) transforms, $\left\{\mathcal{T}_k\right\}_{k=1}^K$. We adopt and extend this approach for sample selection in an active learning setting. The scoring function, $\mathcal{S}$, is the \emph{feature reconstruction error} score, which is the norm of the difference between the original feature vector and the pre-image of its corresponding reduced embedding:
\begin{equation}
    \mathcal{S} \triangleq FRE(z|y=c_k) = \|z-(\mathcal{T}_k^{\dagger} \circ \mathcal{T}_k)(z)\|_2.
\end{equation}
Here, $\mathcal{T}_k$, is the forward PCA transformation, and $\mathcal{T}_k^{\dagger}$ is its Moore-Penrose pseudo-inverse. Note that the transformation used is dependent on the predicted class label. Intuitively, we wish to select samples that are most distant from the reduced-dimension PCA subspace. 


\section{Experiments and Results}
\label{sec:experiments}
\subsection{Experimental setup}

We perform our experiments on image classification task with CIFAR-10~\cite{krizhevsky2009learning}, Fashion-MNIST~\cite{xiao2017fashion} and Street View House Numbers (SVHN)~\cite{netzer2011reading} datasets. We evaluate our proposed methods SCAL, DFM and FeatureSim while comparing with other high-performing methods including Learning loss~\cite{yoo2019learning}, CoreSet~\cite{sener2018active} and Bayesian active learning by disagreement (BALD)~\cite{gal2017deep}, Random and Entropy. 

We use ResNet-18~\cite{he2016deep} model architecture for all the methods and datasets under study. We use the same hyperparameters for all the models for a fair comparison. The models are trained with the SGD optimizer with an initial learning rate of 0.1, momentum of 0.9, and weight decay of 0.0005 for 200 epochs. As part of the learning rate schedule, the initial learning rate is multiplied by 0.1 at epoch 160. 
At each iteration of the active learning cycle, the models are trained with labeled samples acquired through the query strategy from respective methods. At the end of each active learning cycle, the models are evaluated with the labeled test set. We use these hyperparameters for all three datasets in our experiments. 

In our experimental setup, we assume there are no labels available initially in the training set. The initial unlabeled set $\mathcal{D}_{\mathrm{U}}$ has 50K samples for CIFAR-10, 60K samples for Fashion-MNIST and 73.2K samples for SVHN. We set $\mathrm{M}=1000$ for CIFAR-10 and SVHN datasets, and $M=500$ for Fashion-MNIST. As described in Section \ref{sec:methods}, the model is trained iteratively with a sequence of labeled datasets $\left[\mathcal{D}_{\mathrm{L}}^{1}, \mathcal{D}_{\mathrm{L}}^{2}, \dots, \right]$. Furthermore, at every iteration $n$ of the active learning cycle, we obtain a random subset $\mathcal{D}_{\mathrm{S}} \subset \mathcal{D}_\mathrm{U}^n$ of 10K samples from the remaining unlabeled sets, from which M informative samples are selected using query strategy. 
This subset strategy in the unlabeled pool has been suggested in~\cite{beluch2018power,yoo2019learning} to avoid picking overlapping similar samples. We repeat this cycle for 10 iterations in our evaluation. The models are evaluated with independent labeled test samples at the end of each active learning iteration.

\subsection{Evaluation metrics}
We evaluate the models in active learning setup with various metrics{\footnote[1]{Arrows next to each metric indicate lower({$\downarrow$}) or higher({$\uparrow$}) value is better.} including test accuracy ({$\uparrow$}), sampling bias ({$\downarrow$}), expected calibration error ({$\downarrow$})~\cite{naeini2015obtaining}, Brier score ({$\downarrow$})~\cite{brier1950verification} and query time ({$\downarrow$}). We evaluate robustness to out-of-distribution with \textit{area under the receiver operating characteristic curve} (AUROC) ({$\uparrow$})~\cite{davis2006relationship}.}

Sampling Bias measures the class imbalance in the acquired data. We propose the following score which measures the deviation from a balanced (uniform) distribution over the classes:
\begin{equation}
\label{eq:bias}
\text{Sampling Bias} = 1-\frac{\mathcal{H}_{D_L}}{\mathcal{H}_{balanced}}.
\end{equation}
Here, $\mathcal{H}_{D_L}$, is the entropy of the sample distribution over the labeled dataset defined as:
\begin{equation}
\label{eq:bias}
\mathcal{H}_{D_L} = - \sum_{k=1}^K \left(\frac{M_k}{M}\right) \log \left(\frac{M_k}{M}\right),
\end{equation}
where $M_k$ is the number of samples from class $c_k$, and $M=\sum_k M_k$ is the total number of samples. $\mathcal{H}_{balanced}$ is the entropy of a balanced sample distribution for which all $M_k$ are equal. With this definition, sampling bias is always between 0 and 1, with 0 indicating no bias (perfectly balanced sample distribution) and 1 indicating a fully imbalanced distribution in which all samples belong to only one class and the other classes having no samples.

Expected calibration error (ECE)~\cite{naeini2015obtaining} (lower is better) measures difference in the expectation between model confidence and accuracy as defined in Equation~\eqref{eqn:ece}. It is computed after dividing the confidence range [0,1] into S bins. N is the total number of samples and $B_s$ is the set of indices of samples whose prediction confidence falls into the $s^{th}$ bin.
\begin{equation}
\label{eqn:ece}
\mathrm{ECE}=\sum_{s=1}^{S} \frac{\left|B_{s}\right|}{N}\left|\operatorname{acc}\left(B_{s}\right)-\operatorname{conf}\left(B_{s}\right)\right|
\end{equation}

\begin{figure*}[ht]
		\centering     
		\subfigure[Fashion-MNIST]{\includegraphics[width=0.65\columnwidth]{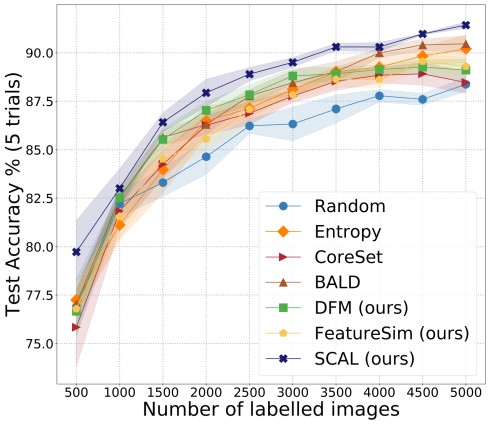}}
		\subfigure[CIFAR-10]{\includegraphics[width=0.65\columnwidth]{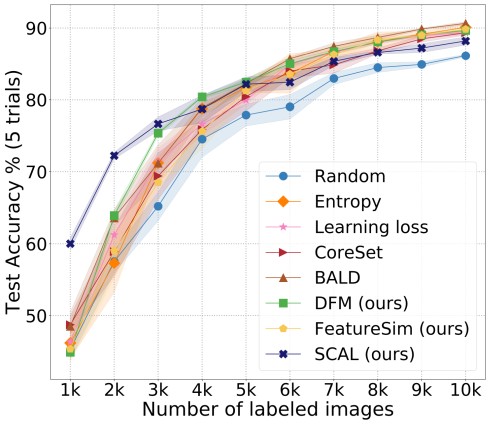}}
		\subfigure[SVHN]{\includegraphics[width=0.65\columnwidth]{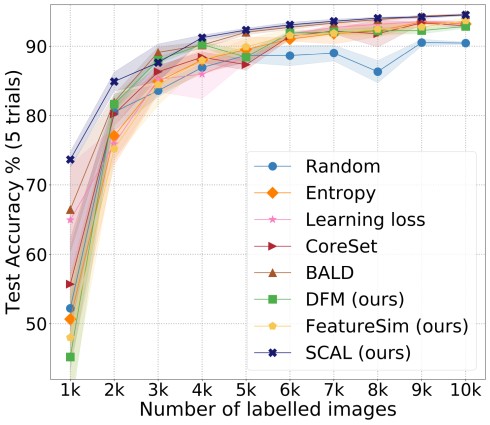}}
		\caption{Test Accuracy evaluation of different query methods in active learning for Fashion-MNIST, CIFAR-10 and SVHN datasets. The shading shows std-dev from 5 independent trials for each method.}
\label{fig:acc}		
\end{figure*}

\begin{figure*}
		\centering     
		\subfigure[Fashion-MNIST]{\includegraphics[width=0.65\columnwidth]{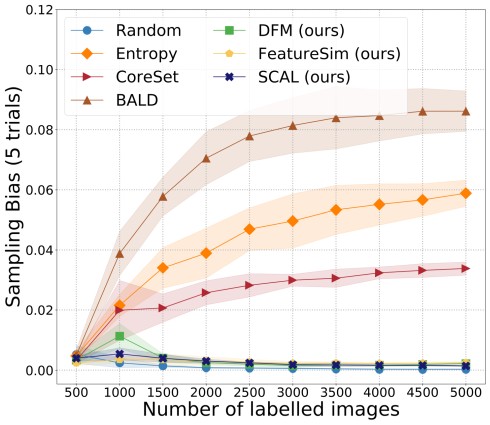}}
		\subfigure[CIFAR-10]{\includegraphics[width=0.65\columnwidth]{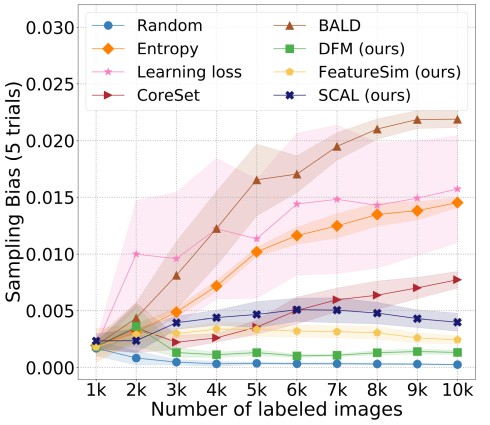}}
		\subfigure[SVHN]{\includegraphics[width=0.65\columnwidth]{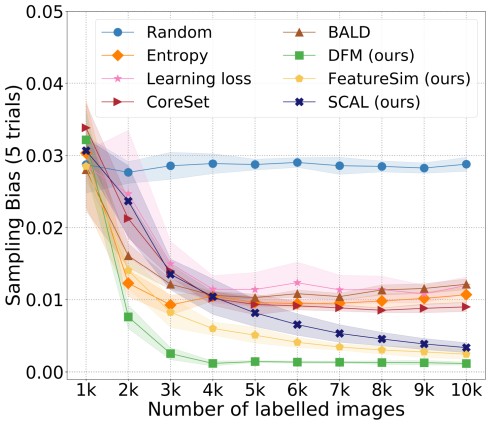}}
		\caption{Sampling bias (lower is better) evaluation of different query methods in active learning for Fashion-MNIST, CIFAR-10 and SVHN datasets. The shading shows std-dev from 5 independent trials for each method. It is to be noted that SVHN has an inherent bias in the dataset with significant imbalance in the number of classes, while Fashion-MNIST and CIFAR-10 are balanced datasets.}
\label{fig:bias}		
\end{figure*}

\subsection{Results}
\label{sec:results}
All the methods shown in Figure~\ref{fig:acc} provide improved accuracy with the increase in training-set size at every active learning cycle and demonstrate better accuracies than the random selection of the samples. SCAL method outperforms other methods in the initial active learning cycles even when fewer training samples are available and yields the best final accuracy for Fashion-MNIST and SVHN datasets. The test accuracy when the model is trained using entire training set with cross entropy (FashionMNIST: 94\%, CIFAR10: 93.04\%, SVHN: 95.93\%), or contrastive (FashionMNIST: 94.43\%, CIFAR10: 92.69\%, SVHN: 95.54\%) loss indicates the upper bound on the accuracy scores. 

\begin{figure*}[ht]
		\centering     
		\subfigure[Fashion-MNIST]{\includegraphics[width=0.65\columnwidth]{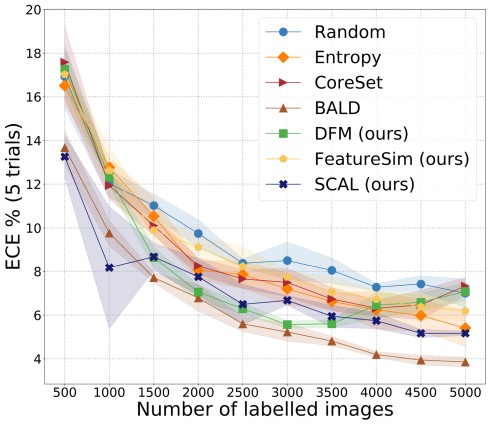}}
		\subfigure[CIFAR-10]{\includegraphics[width=0.65\columnwidth]{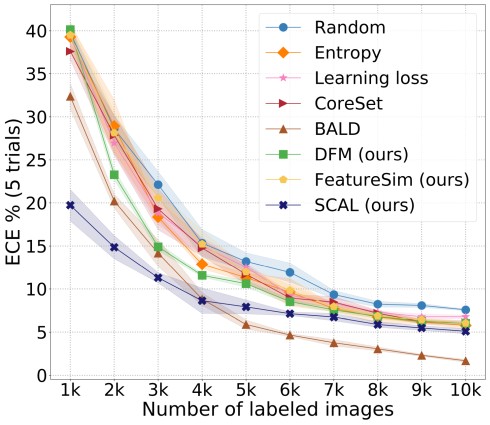}}
		\subfigure[SVHN]{\includegraphics[width=0.65\columnwidth]{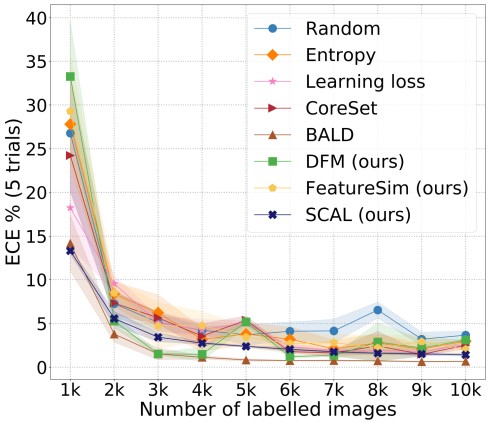}}
		\caption{Expected Calibration Error (ECE) (lower is better) evaluation of different query methods in active learning for Fashion-MNIST, CIFAR-10 and SVHN datasets. The shading shows std-dev from 5 independent trials for each method. Lower ECE is better, indicating the model is well-calibrated. BALD yields the lowest ECE, but it is computationally expensive as noted in Table\eqref{tab:sample-table}.}
\label{fig:ece}		
\end{figure*}

\begin{figure*}[ht]
		\centering     
		\subfigure[ out-of-distribution: AUROC {$\uparrow$} ]{\includegraphics[width=0.65\columnwidth]{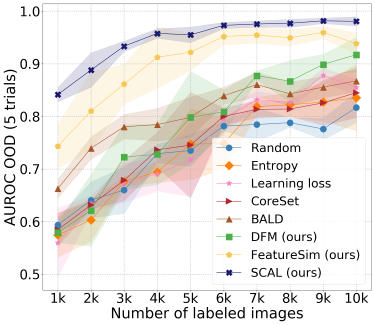}}
		\subfigure[Dataset shift: Accuracy {$\uparrow$}]{\includegraphics[width=0.65\columnwidth]{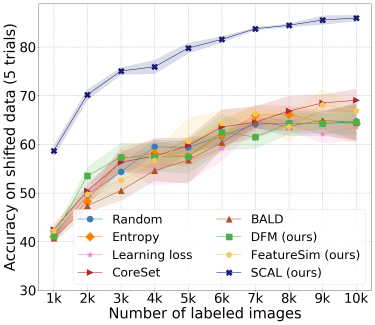}}
		\subfigure[Dataset shift: ECE {$\downarrow$}]{\includegraphics[width=0.65\columnwidth]{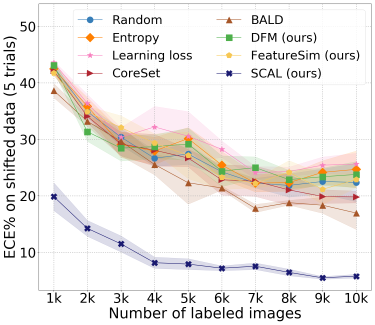}}
		\caption{Robustness to out of distribution and dataset shift (in-distribution: CIFAR-10, out-of-distribution (OOD): SVHN, dataset shift: CIFAR-10 corrupted with Gaussian blur). (a) AUROC of OOD detection shows SCAL outperforms existing methods by more than 10\%. (b) Accuracy on shifted data (Gaussian blur): SCAL is more accurate even under dataset shift due to better feature representation. (c) ECE on shifted data, SCAL yield reliable confidence under dataset shift as reflected by lower ECE. 
		}
\label{fig:robustness}
\end{figure*}

\textbf{Sampling bias:} The plots shown in Figure~\ref{fig:bias} demonstrate the sampling bias across classes in the acquired data. Our proposed methods (DFM, FeatureSim and SCAL) show lower sampling bias at every active learning cycle. SVHN is a class imbalanced dataset and has the inherent bias between classes, which is reflected in the random selection of the samples at every cycle. But, our proposed methods reduces the bias (shown in Figure~\ref{fig:bias}-c) for the imbalanced SVHN dataset while selecting the most informative samples to train the models. In Table~\ref{tab:sample-table}, we compare the sample query time for all the methods to select 1000 new training samples at every active learning cycle. For a fair comparison, these measurements are captured on the same compute machine and experimental settings for all the methods. The average query time is reported as a relative unit with Entropy method as baseline. The query time measures the time needed to compute the respective scores for selecting the most informative samples. Table~\ref{tab:sample-table} shows that our methods are 26x faster than BALD and 11x faster than CoreSet, while achieving state-of-the-art accuracy and reducing the sampling bias.

\begin{table}[t]
\caption{\textbf{Query time {$\downarrow$}} (unit relative to Entropy method as baseline) for computing the scores to select 1K most informative samples from a subset of 10K samples from the unlabeled pool at each iteration of active learning cycle with ResNet-18/CIFAR-10.}
\label{tab:sample-table}
\begin{center}
\begin{small}
\begin{sc}
\begin{tabular}{lcccr}
\toprule
Query Method & Avg. Query Time unit {$\downarrow$} \\
\midrule
Entropy    & 1.0 \\
Learning Loss & 1.803\\
CoreSet & 15.181\\
BALD{\footnotemark[2]}    & 35.693\\
DFM (ours)    & 1.378 \\
FeatureSim (ours)    & 1.322\\
SCAL (ours)      & 1.373 \\
\bottomrule
\end{tabular}
\end{sc}
\end{small}
\end{center}
\vskip -0.2in
\end{table}
\footnotetext[2]{50 stochastic forward passes (Monte Carlo Dropout)}

\textbf{Model Calibration:} In addition to accuracy improvement in active learning, we evaluate the calibration of the models resulting from different query methods in active learning. It is important for the models to provide reliable confidence and uncertainty measures in addition to providing accurate predictions, which is quantified with Expected Calibration Error (ECE) at every active-learning cycle. BALD method (shown in Figure~\ref{fig:ece}) yields lowest ECE, which can be attributed to multiple Monte Carlo stochastic forward passes during inference. But BALD method needs higher query time due to multiple forward-passes as reported in Table~\ref{tab:sample-table}. Our proposed SCAL method outperforms the rest of the approaches yielding well-calibrated models with lower query time. The shading in the plots shows standard deviation in the results from 5 independent trials for each method.

\textbf{Robustness to distributional shift:} We evaluate the robustness to out-of-distribution data and dataset shift~\cite{moreno2012unifying,ovadia2019can} for the models trained under active learning setting. The models need to be robust to distributional shift as the observed data will shift from the training data distribution in real-world settings. Figure~\ref{fig:robustness}-a compares the out-of-distribution (OOD) detection performance for a model trained on CIFAR-10 and evaluated on SVHN. For each of the methods, the same scoring function that was used to select the samples in active learning cycle is used to compute the AUROC with the scores obtained on in-distribution and OOD data. Our proposed methods yield higher AUROC results compared to other high-performing methods, with SCAL outperforming other methods by a big margin. We studied the robustness to dataset shift with CIFAR-10 corrupted with Gaussian Blur~\cite{hendrycks2019benchmarking}. Figure~\ref{fig:robustness} (b) \& (c) show SCAL method outperforms all other methods by a bigger margin by yielding higher test accuracy and lower calibration error demonstrating the robustness under dataset shift. The superior robustness from SCAL method is attributed to the better feature representations.



\section{Conclusion}
\label{sec:discussion}
We proposed simple, computationally less expensive and yet efficient methods to mitigate sampling bias in active learning while achieving state-of-the-art accuracy. We demonstrated our query strategy reduce sampling bias in active learning from both balanced and imbalanced datasets. The proposed methods yield good model calibration just next to Bayesian active learning by disagreement, with a significantly lesser query time for sample acquisition in active learning. We evaluated the robustness to distributional shift of models derived from various state-of-the-art query strategies and training methods in active learning setting. The supervised contrastive active learning outperforms other high-performing active learning methods by a big margin in robustness to out-of-distribution and dataset shift. 


\bibliography{active_learning}
\bibliographystyle{icml2021}


\clearpage

\end{document}